\documentclass[conference]{IEEEtran}

\usepackage{cite}
\usepackage{graphicx}
\usepackage{amsmath}
\usepackage{array}
\usepackage{multirow}
\ifCLASSOPTIONcompsoc
 \usepackage[caption=false,font=normalsize,labelfont=sf,textfont=sf]{subfig}
\else
 \usepackage[caption=false,font=footnotesize]{subfig}
\fi
\usepackage{url}
\usepackage{bm}
\usepackage{tikz}

\newcommand\copyrighttext{%
  \footnotesize \textcopyright 2019 IEEE. Personal use of this material is permitted.
  Permission from IEEE must be obtained for all other uses, in any current or future
  media, including reprinting/republishing this material for advertising or promotional
  purposes, creating new collective works, for resale or redistribution to servers or
  lists, or reuse of any copyrighted component of this work in other works.
}
\newcommand\copyrightnotice{%
\begin{tikzpicture}[remember picture,overlay]
\node[anchor=south,yshift=10pt] at (current page.south) {\fbox{\parbox{\dimexpr\textwidth-\fboxsep-\fboxrule\relax}{\copyrighttext}}};
\end{tikzpicture}%
}

\begin{document}
\title{Lukthung Classification Using Neural Networks on Lyrics and Audios}


%
\author{
    \IEEEauthorblockN{
        Kawisorn Kamtue\IEEEauthorrefmark{1},
        Kasina Euchukanonchai\IEEEauthorrefmark{2},
        Dittaya Wanvarie\IEEEauthorrefmark{3} and
        Naruemon Pratanwanich\IEEEauthorrefmark{4}
    }
    \IEEEauthorblockA{
        \IEEEauthorrefmark{1}\IEEEauthorrefmark{2}Tencent (Thailand)\\
        \IEEEauthorrefmark{3}\IEEEauthorrefmark{4}Department of Mathematics and Computer Science, Faculty of Science, Chulalongkorn University, Thailand
    }
    \IEEEauthorblockA{
        Email:
            \IEEEauthorrefmark{1}kawisorn@tencent.co.th,
            \IEEEauthorrefmark{2}kasina@tencent.co.th,
            \IEEEauthorrefmark{3}Dittaya.W@chula.ac.th,
            \IEEEauthorrefmark{4}Naruemon.P@chula.ac.th
    }
}


\maketitle
\copyrightnotice

\begin{abstract}
Music genre classification is a widely researched topic in music information
retrieval (MIR). Being able to automatically tag genres will benefit music streaming service providers such as JOOX, Apple Music, and Spotify for their content-based recommendation. However, most studies on music classification have been done on western songs which differ from Thai songs. Lukthung, a distinctive and long-established type of Thai music, is one of the most popular music genres in Thailand and has a specific group of listeners. In this paper, we develop neural networks to classify such Lukthung genre from others using both lyrics and audios. Words used in Lukthung songs are particularly poetical, and their musical styles are uniquely composed of traditional Thai instruments. We leverage these two main characteristics by building a lyrics model based on bag-of-words (BoW), and an audio model using a convolutional neural network (CNN) architecture. We then aggregate the intermediate features learned from both models to build a final classifier. Our results show that the proposed three models outperform all of the standard classifiers where the combined model yields the best $F_1$ score of 0.86, allowing Lukthung classification to be applicable to personalized recommendation for Thai audience.
\end{abstract}


%
\IEEEpeerreviewmaketitle

\section{Introduction}
Lukthung, a unique type of music genre, originated from rural communities in Thailand. It is one of the most prominent genres and has a large listener base from farmers and urban working-class people \cite{c13}. Lyrically, the songs contain a wide range of themes, yet often based on Thai country life: rural poverty, romantic love, aesthetic appeal of pastoral scenery, religious belief, traditional culture, and political crisis \cite{c11}. The vocal parts are usually sung with unique country accents and ubiquitous use of vibrato and are harmonized with western instruments (e.g. brass and electronic devices), as well as traditional Thai instruments such as Khene (mouth organ) and Phin (lute).

It is normal to see on many public music streaming platforms such as Youtube that many non-Lukthung playlists contain very few or even none of Lukthung tracks, compared to other genres e.g. Pop, Rock, R\&B which are usually mixed together in the same playlists. This implies that only a small proportion of users listen to both Lukthung and other genres, and non-Lukthung listeners rarely listen to Lukthung songs at all. Therefore, for the purpose of personalized music recommendation in the Thai music industry, identifying Lukthung songs in hundreds of thousands of songs can reduce the chance of mistakenly recommending them to non-Lukthung listeners.

Many musical genre classification methods rely heavily on audio files, either raw waveforms or frequency spectrograms, as predictors.
Previously, traditional approaches focused on hand-crafted feature extraction to be input to classifiers \cite{c3,c18}, while more recent network-based approaches view spectrograms as 2-dimensional images for musical genre classification \cite{c10}.

Lyrics-based music classification is less studied, as it has generally been less successful \cite{c12}. Early approaches designed lyrics features that were comprised of vocabulary, styles, semantics, orientations and, structures for SVM classifiers \cite{c14}. Since modern natural language processing techniques have switched to recurrent neural networks (RNNs), lyrics-based genre classification can also employ similar architectures.
On the other hand, as Lukthung songs often contain dialects and unconventional vocabulary, we will show later that a simpler bag-of-words (BoW) model on our relatively much smaller dataset can also achieve satisfying results.

Recently, both lyrics and audios have been incorporated into classifiers together \cite{c16,c17}. The assumption is that lyrics contain inherent semantics that cannot be captured by audios. Therefore, both features provide information that complements each other. 

Although both audio and lyrics features have been used in musical genre classification, to the best of our knowledge, none of the previous works have been done on Lukthung classification. 
In principle, many existing methods can be adopted directly. However, classifying Lukthung from other genres is challenging in practice. This is because machine learning models, especially deep neural networks, are known to be specific to datasets on which they are trained. Moreover, Lukthung songs, while uniquely special, are sometimes close to several other Thai traditional genres. Proper architecture design and optimal parameter tuning are still required.

In this paper, we build a system that harnesses audio characteristics and lyrics to automatically and effectively classify Lukthung songs from other genres. Since audio and lyrics attributes are intrinsically different, we train two separate models solely from lyrics and audios. Nevertheless, these two features are complementary to a certain degree. Hence, we build a final classifier by aggregating the intermediate features learned from both individual models. Our results show that the proposed models outperform the traditional methods where the combined model that make use of both lyrics and audios gives the best classification accuracy.


\section{Related work}
\label{Sec:literature}

For automatic music tagging tasks including genre classification, the conventional machine learning approaches involve creating hand-crafted features and using them as the input of a classifier.
Rather than the raw waveforms,  early classification models were frequently trained on Mel-frequency cepstrum coefficients (MFCCs) extracted from audio waves for computational efficiency \cite{c1,c2}.
Other hand-designed features were introduced in addition to MFCCs for music tagging \cite{c3}. Like MFCCs, these audio features such as MFCC derivatives and spectral features typically represent the physical or perceived aspects of sounds. Since these derived features are frame-dependent attributes, their statistics such as means and variances are computed across multiple time frames to generate a feature vector per an audio clip. 
However, the main difficulty with hand-crafted features is that task-relevant features are challenging to be designed. Many feature selection techniques have been introduced to tackle this problem \cite{c5}, but the right solution is yet to emerge. 

Recently, deep learning approaches have been widely explored to combine feature extraction and modeling, allowing relevant features to be learned automatically. Following the success in speech recognition \cite{c6}, deep neural networks (DNNs) have been currently used for audio data analysis \cite{c7,c8,c9}. The hidden layers in DNNs can be interpreted as representative features underlying the audio. Without requiring hand-crafted features from the audio spectrograms, a neural network can automatically learn high-level representations while classification being trained. Instead, the only requirement is to determine the network architecture e.g. the number of nodes in hidden layers, the number of filters (kernels) and the filter size in convolution layers such that meaningful features can be captured. 

In \cite{c10}, several filters different in sizes were explicitly designed to capture important patterns in two-dimensional spectrograms derived from audio files.
Unlike the filters used in standard image classification, they introduced vertical filters lying along the frequency axis to capture pitch-invariant timbral features simultaneously with long horizontal filters to learn such time-invariant attributes as tempos and rhythms. The underlying intuition is that the musical features of interest residing in a 2D spectrogram are not spatially invariant along the frequency and time axes. Accordingly, the common symmetric filters are not effective for feature extraction on spectrograms.


Lyrics-based models for music genre classification are similar to those used in text classification. \cite{c14} categorized lyrics features into five main classes: vocabulary, style, semantics, orientation and song structure, and used them for music genre classification. With the recent success in recurrent neural networks (RNNs) in natural language processing, a modern approach used Hierarchical attention networks (HAN) for music genre classification \cite{c15}. This advance in deep learning methods allows us to encapsulate both meanings and the structure of lyrics. However, this model relied on a large lyrics corpus to train word embedding, which is not practical for small datasets.

\section{Dataset}

\begin{center}
\begin{figure}
  \centering
    \includegraphics[width=\linewidth]{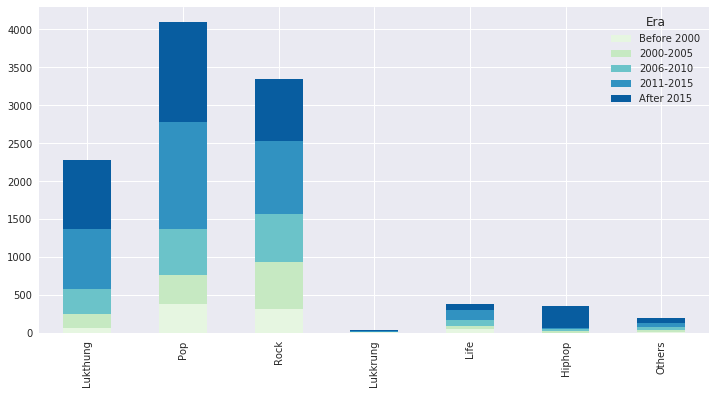}
    \caption{Number of instances in each genre and era classes}
  \label{fig:genre_year_plot}
\end{figure}
\end{center}

We collected 10,547 Thai songs dated from the year 1985 to 2019 where both lyrics and audios were available. The genre of each song, along with other tags related to mood, tempo, musical instruments, to name a few, was tagged by music experts. Genre annotations are Pop, Rock, Lukthung (Thai rural style), Lukkrung (Thai urban style in the 30s to 60s eras), Hip Hop, and Phuea Chiwit (translated as songs for life) which we will refer as Life for the rest of this paper. Here, we are interested in distinguishing Lukthung songs from the others. Fig. \ref{fig:genre_year_plot} shows the number of songs labeled in each genre and era. 

We found that Lukthung songs occupy approximately 20\% of the entire dataset. There are few songs before 2000s. The genre Others in Fig. \ref{fig:genre_year_plot} refers to less popular genres such as Reggae and Ska.
Since the era also affects the song style, the model performance may vary. We will also discuss the results based on different eras in Section \ref{Sec:experiments}.

\section{Data preprocessing}
\label{Sec:preprocessing}
Our goal is to build machine learning models from both lyrics and audio inputs. 
We describe the preprocessing steps carried out in our data preparation for each data type below.

\begin{figure*}
  \centering
    \includegraphics[width=\linewidth]{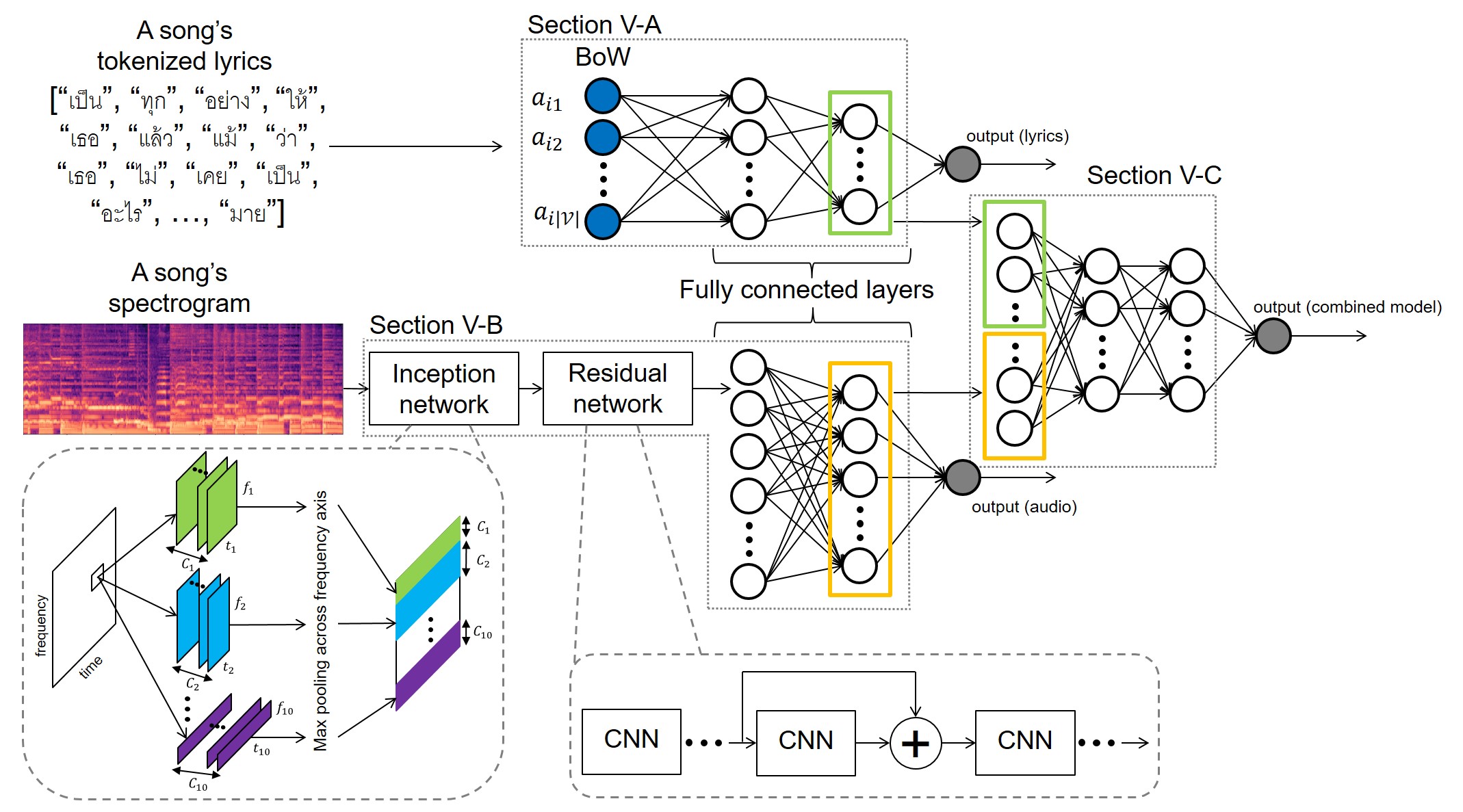}
    \caption{Overall model architecture}
  \label{fig:overallarch}
\end{figure*}

\subsection{Lyrics input}
\label{Sec:lyricfeature}
We constructed word-based features using the entire lyrics from the beginning to the end of the songs. The lyrics were firstly tokenized. We tried several Thai word tokenizers and chose the \texttt{deepcut} library  \footnote{\url{https://github.com/rkcosmos/deepcut}} due to its best performance on this task. It is noted that the artists’ names were not included, allowing the model to perform classification on songs from unknown or never-seen-before artists.
An example of tokenized lyrics is shown in Fig. \ref{fig:overallarch} (top-left), which is passed through the lyrics model described in Section \ref{Sec:lyricmodel}.

\subsection{Audio input}
\label{Sec:audiofeature}
For each song, we excerpted a 10-second clip from an audio file in its chorus part. We used the chorus part solely not only for computational reasons but also because it usually contains all musical instruments present in the song, hence containing the richest information. We approximated that the chorus part came after 30\% of the total song duration from the start. Finally, we extracted the Mel spectrograms from the excerpted audio clips with the following specifications:

\begin{itemize}
    \item Sampling rate: 22050 Hz i.e. frame size = 4.53e-5 s
    \item Frequency range: 300-8000 Hz
    \item Number of Mel bins: 128
    \item Window of length (n\_fft): 2048
    \item Time advance between frames (hop size): 512
\end{itemize}

\begin{figure}
  \centering
    \includegraphics[width=\linewidth]{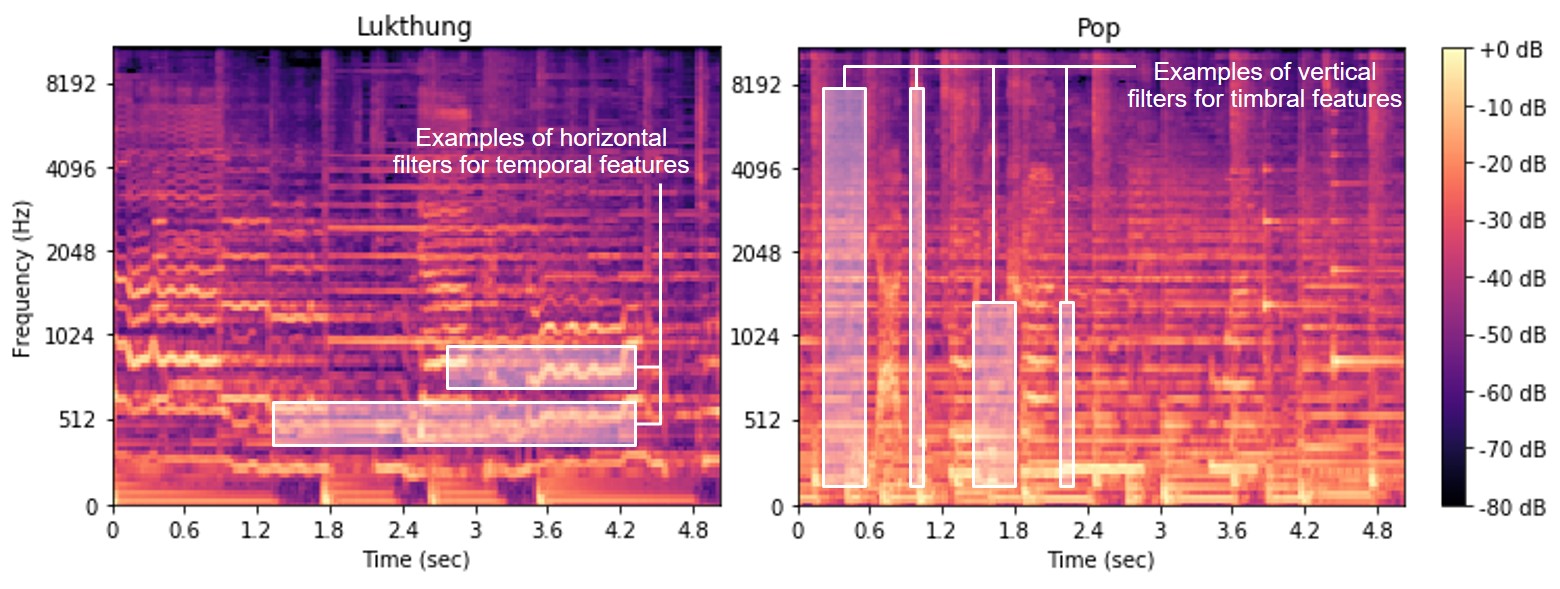}
    \caption{Mel spectrograms (shown only half of the input time length) along with vertical and horizontal filters}
  \label{fig:lukthung_pop_filters}
\end{figure}




Examples of extracted spectrograms are demonstrated in Fig. \ref{fig:overallarch} (center-left) and Fig. \ref{fig:lukthung_pop_filters}.
We may see from Fig. \ref{fig:lukthung_pop_filters} that audio clips from different genres have different characteristics. 
We will discuss the audio model that captures such features in Section \ref{Sec:audiomodel}.

\section{Proposed models}
\label{Sec:model}
We propose separate models for the lyrics and audio features. We also propose the combined model that aggregates the intermediate features from both models.

\subsection{Lyrics Model: BoW-MLP}
\label{Sec:lyricmodel}
Since most Lukthung songs have a different set of vocabulary compared to other genres, we propose a simple bag-of-words (BoW) model. The vocabulary $\mathcal{V}$ is constructed using a larger set of unlabeled and labeled lyrics, from roughly 85k songs. We filter out words that are longer than 20 characters and words that appear less than 10 times in the entire lyrics corpus. The lyrics of each song is represented by a normalized bag-of-words vector $\bm{a}_i$ using the vocabulary $\mathcal{V}$. Let $c_{i,j}$ denote the number of occurrences of word $j$ in the lyrics of song $i$. The normalized count of this word, $a_{i,j}$, is computed as in \eqref{eq:normalized_count}.

\begin{equation}
    a_{i,j} = \frac{\log c_{i,j}}{max \log \bm{c}_{i}} 
    \label{eq:normalized_count}
\end{equation}

The logarithm transformation is applied to smooth the discrete count values. We normalize by the maximum word count within the lyrics of each song because it preserves the sparsity of the BoW features while scaling each feature value to between 0 and 1, inclusive. The BoW is then fed into a two-layer fully connected multi-layer perceptron (MLP). The input layer is comprised of $|\mathcal{V}|$ nodes where $|\mathcal{V}|$ is the vocabulary size, followed by 100 hidden nodes in each intermediate layer before connecting to a single neuron in the output layer. We put rectified linear unit (ReLU) activation functions on the hidden nodes to allow the model to learn non-linear mapping, and place a sigmoid activation function on the output node to obtain the probability whether a given song is Lukthung. The graphical model architecture is depicted in the Fig.\ref{fig:overallarch} (top).

\subsection{Audio Model: Spectro-CNN}
\label{Sec:audiomodel}
We develop the same CNN-based model as stated in \cite{c10} with some architectural modifications to be suited in our data set as elaborated below. The overview of the model is illustrated in Fig. \ref{fig:overallarch} (bottom).

The model is aimed to automatically learn timbral and temporal properties for the genre classification task.
Overall, the model starts with the input layer. These inputs are passed through an inception network to extract features, followed by a residual network. Then, the output from the residual network is fed to a fully connected layers to predict a binary output. 

For each audio file, we construct a 2-dimensional spectrogram spanning across 431 time frames with 128 Mel frequency bins, as described earlier in Section \ref{Sec:audiofeature}. The spectrogram inputs are then put forward through the feature extraction layers where both timbral and temporal features are extracted in parallel using a set of convolutional filters in different dimensions.


Along the frequency axis on the spectrograms, the timbral elements are represented by a stack of bright lines with different degrees of intensity, which involve perceived vocal and tone-related sound. In addition to fundamental frequencies, Lukthung songs often contain Thai traditional musical instruments and vocals that produce overtone partials (either multiples of the fundamental frequencies or any higher constituents of frequencies). To detect such timbral features located along the frequency axis of the spectrograms, we apply vertical filters with variable heights of mel bins across short time intervals, where the taller vertical filters aim to capture the higher overtones. Following the parallel convolution layer, a max pooling is performed across the frequency axis. Examples of vertical filters are highlighted on the left part of Fig \ref{fig:lukthung_pop_filters}.

In parallel with the timbral feature extraction, we place horizontal blocks of filters to learn the temporal elements including tempos and rhythms which can be detected by a drastic change in energy along the time axis.
We handle songs with different tempos and rhythms using variable lengths of filters along the time axis. We refer to such filters as horizontal filters as depicted on the right part of Fig. \ref{fig:lukthung_pop_filters}.
Unlike the architecture in \cite{c10} where a 1D convolution was performed across the time axis after the spectrograms were mean-pooled along the frequency axis, we expand the horizontal filters to cover a small frequency bins and apply them on the inputs before a max pooling layer.
This is because we aim to preserve vibrato or coloratura, a remarkable trait presented in Lukthung singing style.
Vibrato is a characteristic of sound oscillating within a small range of frequencies over time, considered a hybrid timbral-temporal features. 
An example of vibrato is the wavy bright lines in the spectrogram illustrated in the left part of Fig. \ref{fig:lukthung_pop_filters}. In contrast, spectrograms of Pop songs, as depicted in the right part of Fig. \ref{fig:lukthung_pop_filters}, have only straight lines.
We reason that this type of feature might vanish if we perform the average pooling process directly on the spectrogram inputs as done in \cite{c10}.
Table \ref{tab:filters} summarized all filter sizes and the number of filters used in our feature extraction module.

\begin{table}
    \centering
    \caption{Summary of filters used in our feature extraction module}
    \label{tab:filters}
    \begin{tabular}{||c|c|c|c||}
        \hline
        Feature& Frequency & Time & \multirow{2}{*}{Number of filters}  \\
        type & (Mel bins) & (units) & \\
        \hline\hline
        \multirow{6}{*}{Timbral} & 115 & 7 & 32  \\
         & 115 & 3 & 64 \\
         & 115 & 1 & 128  \\
        & 51 & 7 & 32  \\
        & 51 & 3 & 64 \\
         & 51 & 1 & 128 \\
        \hline\hline
       \multirow{4}{*}{Temporal}  & 7 & 32 & 32 \\
         & 7 & 64 & 32 \\
         & 7 & 128 & 32 \\
         & 7 & 165 & 32 \\
         \hline
    \end{tabular}
\end{table}

After obtaining the representations of both timbral and temporal features, we concatenate them along the frequency axis and pass them into the binary classification module to predict whether a given audio clip is considered a Lukthung genre.
Within the classification module, a residual network with three convolution layers followed by a fully connected network is implemented.
Details of the residual network architecture can be found in \cite{c19}.
Briefly, a residual network is stacked of convolution layers with alternative connections that skip from one layer to the next. Such bypass connections mitigate the effect of gradient vanishing and importantly allow the model to learn to identity functions, ensuring that the higher layers perform at least as well as the lower layers.

\subsection{Combined model}
\label{Sec:combinedmodel}

Since both lyrics and audios carry rich information about the Lukthung genre, we decide to combine both data to perform the classification task. 
Instead of using the single predicted probabilities from each model, we extract the learned representations from the last layer of both models and concatenate them to form a new feature vector.
We reason that some features extracted from one type of inputs are complementary to the other and should be learned simultaneously.
Based on these pre-trained features, we construct an additional feed-forward neural network comprising 800 nodes (100 from the lyrics model and 700 from the audio model) in the input layer, fully connected with 2 layers with ReLU activation and a single output node for the binary classification.
The process is illustrated in the right part of Fig. \ref{fig:overallarch}.


\section{Results}
\label{Sec:experiments}

We randomly split the dataset into a training set, validation set, and test set using the ratio of 0.55:0.2:0.25. 
We trained our models, BoW-MLP, Spectro-CNN, Combined, and all baselines, on INTEL Xeon CPU E5-1650  v3 3.50 GHz with 12 cores, 80GB RAM, and a GeForce GTX 1080 GPU. 

For comparison, we built the following baseline classifiers on the audio inputs using default parameters implemented in the Scikit-learn library\footnote{\url{https://scikit-learn.org/stable/}}.
Using the MFCCs derived from the audio files as inputs, a simple logistic regression (LR), a tree-based random forest (RF), and support vector machines with a linear kernel (SVM-Lin) and a polynomial kernel (SVM-Poly) were trained.

\begin{center}
\begin{table}
    \centering
    \caption{$Precision$, $recall$, and $F_1$ score on test dataset}
    \label{tab:result_test}
    \begin{tabular}{||c | c | c | c | c||} 
 \hline
Input  & Model & Precision & Recall & $F_1$  \\
 
 \hline\hline
 
Lyrics & \textbf{BoW-MLP}  & 0.8581 & 0.7631 & 0.7905 \\ 
 \hline\hline 
\multirow{5}{*}{Audios} & RF & \textbf{0.9500} & 0.1313 & 0.2308  \\ 
 & LR  & 0.5263 & 0.5760 & 0.5501  \\ 
 & SVM-Lin  & 0.6939 & 0.0783 & 0.1408  \\ 
 & SVM-Poly  & 0.5574 & 0.5369 & 0.5469  \\ 
  & \textbf{Spectro-CNN}  & 0.8494 & 0.7397 & 0.7730 \\ 
 \hline\hline

 
 Lyrics $\&$ & \multirow{2}{*}{\textbf{Combined model}}  & \multirow{2}{*}{0.8996} & \multirow{2}{*}{\textbf{0.8344}}  & \multirow{2}{*}{\textbf{0.8561}}\\ 
 Audios & & & & \\
 \hline
\end{tabular}

\end{table}
\end{center}

We evaluated the models based on precision, recall, and $F_1$ scores for handling the imbalanced data. 
The results are shown in Table \ref{tab:result_test} where our proposed models and the best performing scores for each measure are highlighted in bold. Despite the best precision, RF performed unacceptably poor to recall Lukthung songs, yielding an extremely low $F_1$ score.
In contrast, our models trained solely on either lyrics or audios already had significantly higher performance in both recall and precision than other traditional models.
The best classifier based on $F_1$ scores lies on the combined model, supporting our hypothesis that lyrics and audio features are complementary determinants for Lukthung classification.

\begin{figure}
  \centering
    \includegraphics[width=\linewidth]{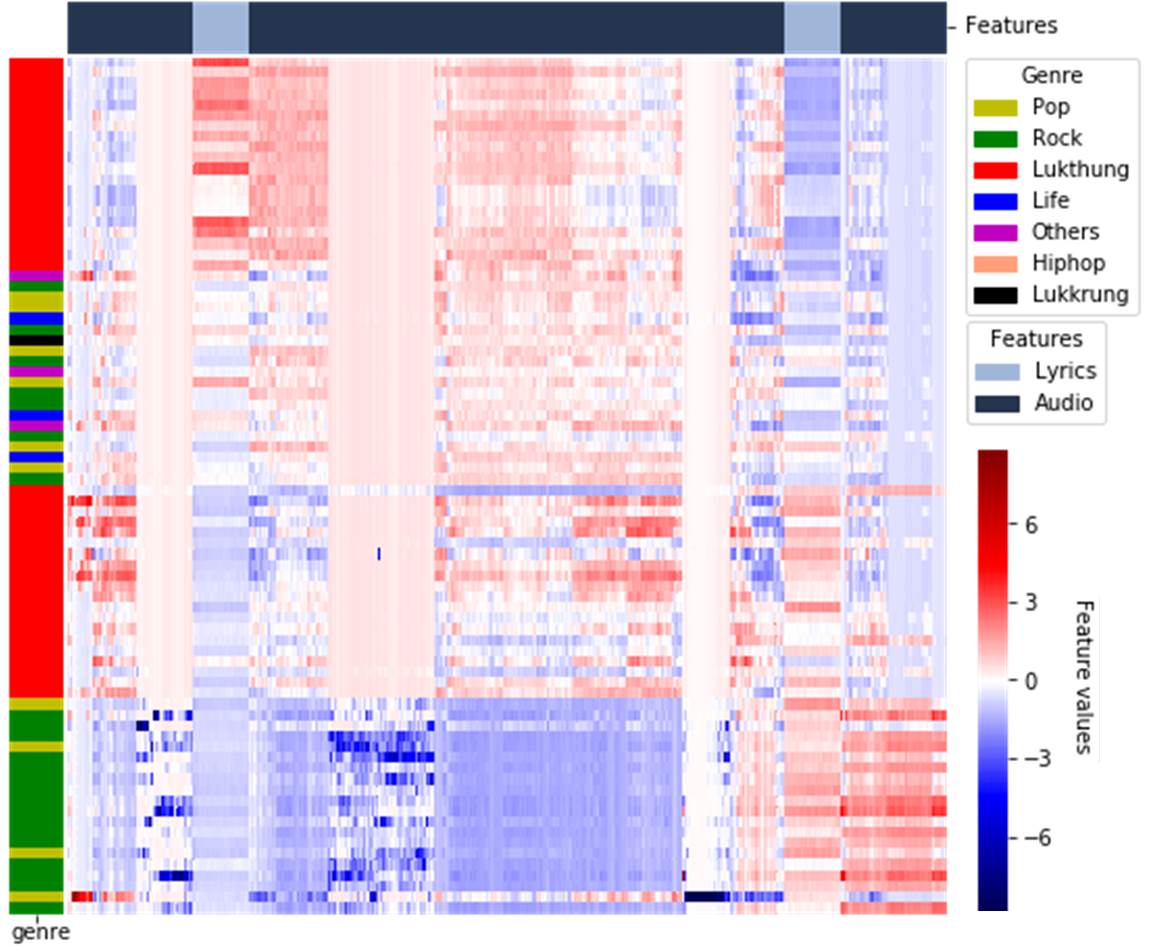}
    \caption{Feature values of selected songs, taken from the input layer of the combined model}
  \label{fig:featureHeatmap}
\end{figure}


We selected the top-20 songs with high predictive confidence to perform feature analysis in these categories: true positive (TP), false positive (FP), false negative (FN), and true negative classes (TN). As an input in the combined model, the learned representations in the last intermediate layer prior to the output node from the lyrics and audio models were extracted and examined below.

Fig. \ref{fig:featureHeatmap} shows such feature values of the selected songs. Each column represents a feature, while rows are songs grouped by the categories mentioned above, respectively. We also clustered the features values (columns) so that similar values were put together. The columns in light blue correspond to the lyrics (word) features, whereas those in dark blue represent the audio features. Note that the combined model carried 800 nodes, 100 word features and 700 audio features, in the input layer.

The heatmap clearly divides the lyrics features into two groups which were exploited differently by the model. Songs predicted as Lukthung had positive values on the first group of lyrics features and negative values on the other, and vice versa for songs predicted as non-Lukthung.
We can see that lyrics features are similar within the predicted classes, separating the top (TP and FP) half from the bottom (FN and TN) half. 

We inspected the false positive songs and found that some of them were Life and Lukkrung as well as Lukthung songs sung with Pop/Rock artists. The extracted lyrics and audio features from these genres are typically similar to Lukthung by nature.

Having scrutinized the list of false negative songs, we categorized them into two sets. The first type is recent Lukthung songs whose lyrics are composed of the standard Thai dialect but may be sung using a non-standard accent with no or only few vibratos. Our audio features hardly capture such accent. Moreover, the musical instruments played in this class of songs are more similar to Pop/Rock songs. This phenomemon is common in songs from new eras. The other type is non-Lukthung songs incorrectly labelled as Lukthung in the dataset, or non-Lukthung songs sung by Lukthung artists.

On the other hand, the values of audio features in the true negative class are much different from other classes. These songs were mainly labelled as Pop and Rock. This indicates that, the audio features of non-Lukthung songs, in general, considerably differ from Lukthung songs. 

\begin{figure}
    \centering
    \includegraphics[width=\linewidth]{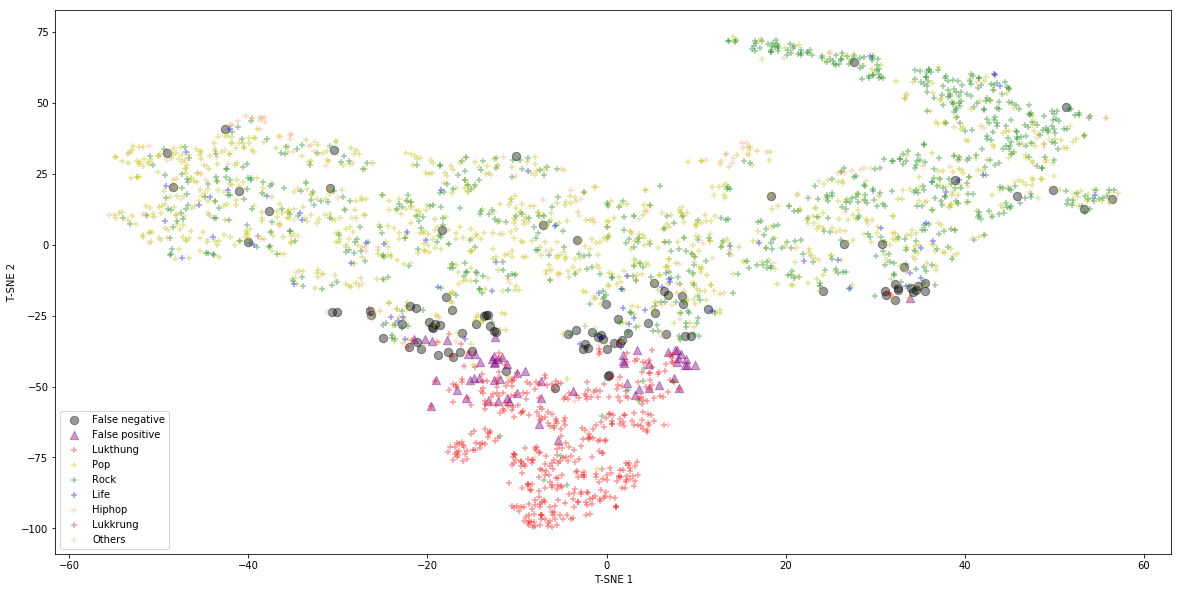}
    \caption{Feature embeddings from the last hidden layer in the combined model}
    \label{fig:tsne_embeddings}
\end{figure}

To visualize the effectiveness of the learned features, we extracted the last hidden layer of the combined model and plotted them on a lower dimensional space using t-SNE as shown in Fig. \ref{fig:tsne_embeddings}. We can clearly see Lukthung and non-Lukthung songs are substantially separated, implying that the features were well extracted by the lyrics and audio models. 
Most of the falsely predicted songs are on the boundary between Lukthung and non-Lukthung songs.
While the false positive songs form a single cloud at the boundary, the Lukthung songs classifed as non-Lukthung (false negatives) are scattered over the non-Lukthung space. This dispersion supports our previous explanation that some Lukthung songs are similar to Pop/Rock songs and moreover not limited to just one group of them.

\section{Conclusion}
\label{Sec:conclusion}

In this paper, we have presented novel classification models to identify Lukthung music from other genres using lyrics and audio inputs. 
Due to a unique set of vocabulary usually used in Lukthung songs, a bag of words representation together with a simple neural network with a few hidden layers is sufficient to distinguish Lukthung from non-Lukthung songs.
The audio inputs, on the other hand, require a more sophisticated model to find patterns across frequency bins and time intervals.
Our approach applies multiple filters on the raw audio spectrograms to automatically learn different types of features such as overtones, tempos, and vibratos.
These abstract features are used later for classification using a residual network with skip connections in deep networks.
Using each input type individually yield satisfying results, outperforming all of the standard classifiers.
Moreover, we show that extracting the pre-trained features from both models and combining them substantially improve the overall performance for Lukthung classification. 

Country songs, which includes Lukthung, Lukkrung, Life and Mor-lam, bear some resemblance to each other in the distributions of words used in lyrics.
This problem may be tackled with document-level, instead of word-level, representation such as semantic word vectors together with such sequence models as recurrent neural network.
With more exposure to contemporary culture, some modern Lukthung songs are now adopting musical instruments and several sound techniques in close proximity to non-Lukthung songs in the old days. However, 
vocals might serve as the main remaining determinant that makes Lukthung differentiable from other genres. 
Thus, isolating singing voice from instrumental and designing vocal-specific filters may beneficially improve the classification outcomes. One example of voice-specific features is to capture the accent of the singer.

Our approach for Lukthung classification can effectively accommodate personalized music recommendation.
Using our model, the system can classify streaming songs and automatically generate a comprehensive list of Lukthung songs in preparation for further music suggestion.
Additionally, further analysis on the features extracted from the models can advance our understanding on how Lukthung songs evolve over eras.

\bibliographystyle{IEEEtran}

\bibliography{lukthungbib}
%



\end{document}